# Intelligent Vision Based Wear Forecasting on Surfaces of Machine Tool Elements

Tobias Schlagenhauf*, Niklas Burghardt

Karlsruhe Institute of Technology
wbk Institute of Production Science, Karlsruhe, Germany

*Abstract*— To realize autonomous production machines it is necessary that machines are able to automatically and autonomously predict their condition. Although many classical as well as Deep Learning based approaches have shown the ability to classify faults, so far there are no approaches that go beyond the basic detection of faults. A complete, image based predictive maintenance approach for machine tool components has to the best of our knowledge not been investigated so far. In this paper it is shown how defects on a Ball Screw Drive (BSD) can be automatically detected by using a machine learning based detection module, quantified by using an intelligent defect quantification module and finally forecasted by a prognosis module in a combined approach. To optimize the presented method, it is shown how existing domain knowledge can be formalized in an expert system and combined with the predictions of the machine learning model to aid quality of the prediction system. This enables the practitioner to forecast the severity of failures on BSD and prevent machine breakdowns. The work is intended to set new accents for the development of practical predictive maintenance systems and bridging the fields of machine learning and production engineering.

The code is available under:
https://github.com/2Obe/Pitting_Pred_Maintenance

## I. Introduction

In the course of the Industrial Internet of Things (IIoT), fully automated machines are becoming increasingly important (Fink et al. 2020). One of the main goals of IIoT is to realize batch size one through a fully automated production (Jeschke et al. 2017). Unexpected machine failures can cause massive delays in the supply chain in just-in-time production and reduce the overall equipment effectiveness of machines. To act with maintenance measures before the failure of a machine occurs, it is necessary to predict failures on machine tool components at an early stage. The goal of condition monitoring systems is to record the current condition of a machine. Subsequently, conclusions are drawn from the data obtained to forecast failures and predict the remaining time till maintenance is needed, also known as predictive maintenance (Wickern 2019). Enabling machines to find defects by themselves and make decisions regarding their wear condition is an enabler to find the sweet spot to maximize the overall equipment effectiveness without the risk of sudden and unforeseen machine breakage due to worn parts (Jeschke et al. 2017). The automation of the maintenance process consists of two critical steps where the first step is to accurately detect failures respectively evaluate the actual health of a component. In the second step, the current condition has to be forecasted to plan maintenance operations in time. The authors show the implementation of a novel autonomous and vision-based condition monitoring system using the example of the ball screw drive (BSD) to then forecast the current failure state using an automatic regression approach. The BSD is chosen as a component since the BSD is one of the most important machine tool components in various industries and studies showed that the BSD is one of the main responsible components for machine tool failures (Schopp 2009). The rolling elements cause material fatigue on the spindle over time which finally results in small breakouts on the spindle's surface called pitting (Fig. 2). These breakouts could also happen on the rolling elements as well, though this will lead to defects on the spindle later on. Surface defects indicate wear but do not lead to a sudden machine breakdown (Klein 2011) which was further validated in life time experiments described later on.

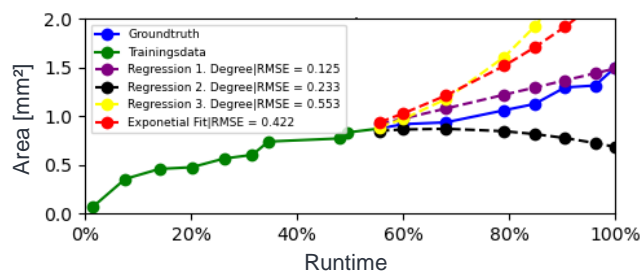

Fig. 1 Defect Size Prediction of different candidate functions

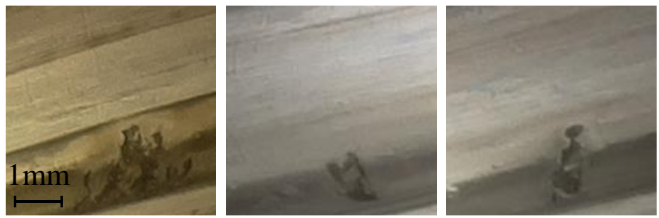

Fig. 2 Examples of pittings on the spindle. All images have the same original size and ration hence the scale bar and the ratios are easily transferable to all other images.

Though, after a sufficient operation time has passed, these pittings influence the performance and finally can lead to the breakdown of a machine. The progress of these failures depends on various conditions such as the load, contaminations as well as the lubrication strategy (Forstmann 2010). It has been shown that the actual lifetime of the BSD can vary largely from the nominal lifetime (Münzing 2017; Spohrer 2019). For this reason, a universal definition over all industries of fixed time limits for the replacement of the BSD is not appropriate and has to be defined by the users based on the acceptable wear for a specific process. This makes the ball screw drive a perfect example for condition monitoring and predictive maintenance. In the here presented work, the authors describe an approach with which image data obtained from a camera system as described in (Schlagenhauf et al. 2019) can be automatically examined for the occurrence of pittings. The size of the pittings is extracted and predicted using a regression-based wear forecaster (Fig. 1). Depending on the user's definition of permissible wear on the component, the BSD can be replaced in time before the mechanic breakdown of the component.

As the main contribution of this paper, the authors provide a novel, vision-based end to end predictive maintenance system from the detection of the failures over the quantification of the size to the prognosis of the future failure severity.

The paper is structured as follows: In section two, an overview on condition monitoring and predictive maintenance systems is given where the focus lies on vision-based systems in the context of machine tools. The third section describes the own approach where the authors start by giving an overview and describing the data used for model building. Next, the proposed predictive maintenance system, together with its sub-components, is described. The authors elaborate the pipeline and show the results of the sub-components in direct connection. The fourth section contains a validation section where the model is validated on new data. The fifth section concludes the paper and states open research questions before it ends with an acknowledgement.

## II. RELATED WORK

The field of condition monitoring and predictive maintenance in the industry is vast which is why the authors want to give a quick overview and further concentrate on the field of machine learning based vision systems to monitor metallic surfaces. Second, the focus should be laid on prognostic systems predicting the progress of failures for predictive maintenance purposes.

Approaches trying to monitor the current state of machines, tools and machine tool components often use the signals of motor current, vibration, temperature and acoustic emission. Exhaustive reviews can be found in (Mohanraj et al. 2020; Iliyas Ahmad et al. 2020; Vanraj et al. 2016; Goyal et al. 2018). Most of the former approaches use a mixture of classical signal processing techniques, where the more recent approaches combine the signal processing techniques with machine learning and deep learning approaches. The field of vision-based condition monitoring techniques is smaller and in the field of machine tools mainly focused on monitoring the wear of cutting tools. (Lutz et al. 2020) implement a new automatic machine learning setup where they remedy the tedious task of hyper-parameter set up for machine learning systems by applying amongst other, different data augmentation techniques to build robust classifiers for the detection of wear on cutting tools. (Wu et al. 2019) use classical CNNs to predict the wear on classical cutting tools. (Sun et al. 2020) try to visualize the condition of a machine by applying machine learning techniques and creating heat maps from the machine components to indicate noticeable areas. (Janssens et al. 2018) for instance, follow a similar approach by combining thermal images of motors and machines with deep convolutional neural networks to predict failures for instance in bearings which result in an anomal temperature increase in specific parts of the component. (Mikołajczyk et al. 2017) use a hybrid approach to detect wear on cutting tools by combining a threshold-based approach as kind of a first filter with a neural network which is then trained on the processed pixels. (Cha et al. 2018) use a quasi-real-time CNN-based approach to detect different types of damages on different metallic and concrete surfaces. (Sadoughi and Hu 2019) combine a vision based convolutional neural network approach with the physical knowledge of wear progression on roller bearings to detect failures. Other approaches focus on detecting defects on metallic surfaces, not necessarily in the context of machine tools. (Dutta et al. 2018) use an indirect but reliable process of measuring tool wear by evaluating the surface quality of the produced products. (La et al. 2018) present an automatic climbing robot prepared with a machine vision system to automatically climb bridge structures and detect cracks. (Dong et al. 2015) use processed Shearlets to classify textures like they could also be observed on technical surfaces like steel, concrete, wood or fabric. (Chu et al. 2017) use a so-called quantile net to detect different defects on metallic surfaces. (Ghorai et al. 2013) present a somewhat earlier approach for the detection of defects on hot-rolled steel products which represents a better studied subfield for automatic flaw detection on metallic surfaces. A newer approach where the promising architecture of Siamese-networks, like it is used by the authors in (Schlagenhauf et al. 2020), is used to classify surface defects, can be found in (Deshpande et al. 2020). Here (Deshpande et al. 2020) use an data efficient one-shot learning approach to recognize manufacturing defects on steel surfaces. (Luo et al. 2020) present a general review on the subfield of defect recognition on steel products, hot- and cold-rolled steel strips. Another interesting approach which implements deep learning techniques for the detection of defects on the NEU dataset (Song and Yunhui 2019) can be found in (He et al. 2020) where they use convolutional neural networks to produce features which are then combined in multilevel feature fusion networks. A somewhat older, general review of vision-based approaches to detect defects on steel surfaces can be found in (Neogi et al. 2014). Though, the field of vision-based condition monitoring specifically for the supervision of machine tools and machine tool components is rare. Related approaches like on rail surfaces (Faghih-Roohi et al. 2016) and concrete structures (Koch et al. 2015) implementing similar machine learning based techniques as for the detection of defects on metallic surfaces with the entitled goal for specifically monitoring the condition of the mentioned structures are rare examples in the literature. Based on the recognition of an anomaly in the operating behaviour of a machine tool, not only limited to vision-based approaches, the literature shows some works which try to forecast the detected signal to implement some sort of predictive maintenance. (Mihajlov et al. 2020) present an approach based on a vibro-diagnostic model for predictive maintenance of rotary machines. (Dias et al. 2020) present a machine learning based approach to classify the vibration signal of a machine into normal or anomalous. (Carvalho et al. 2019) provide a systematic literature review on the field of machine learning methods applied for predictive maintenance.

It is notable that many approaches use the terminology of predictive maintenance though strictly speaking they are condition monitoring systems since they solely decide if a machine or a machine tool is working properly or not. The necessary part of prediction into the future is often not clear. Hence, it can be summarized that there are limited end to end predictive maintenance systems in general, and no systems which operate on the image based surface characteristics of a failure on metallic surfaces in general and on BSD in specific – implementing Machine Learning techniques. The reason for this can be found in the fact that building such models necessitates proper datasets showing the evolution of defects, which are, to the best of the author's knowledge, not available so far.

### III. OWN APPROACH & RESULTS

This section describes the different modules of the BSD predictive maintenance system which are depicted in Fig. 4. The first part is the raw data, generated in lifetime experiments using the camera system described in (Schlagenhauf et al. 2019). The generated dataset has the specific property that it shows a continuous progression of failures and hence depicts the whole wear history from the start of operation until the mechanical breakdown of the component. On this dataset a so called Pitting Detection model is build which is able to return a bounding box approximating the size of a defect. This result can be further refined by using a new approach for the calculation of the defect area based on a combination of a classical threshold-based method in combination with a convolutional neural network (CNN) predicting the threshold for the calculation of the area with the threshold model. The extracted failure area is then processed in a forecasting module which is trained on historical defect data. The model can finally be used to predict the future area of the size of new failures.

### A. Data Set

Using a BSD test bench to artificially wear ball screw drives, the authors generated images of pittings in a temporal relationship. Early images show no or only small pittings which then grow over time until the component fails. A progression of pittings is depicted in Fig. 14. As stopping criteria for the experiments the authors defined the mechanical breakdown of the system. During the experiments the authors mounted a camera system as described in detail in (Schlagenhauf et al. 2019) close to the nut of the ball screw drive such that the system looks radially onto the spindle and returns images of the surface of the spindle. The experimental setup is depicted in Fig. 3.

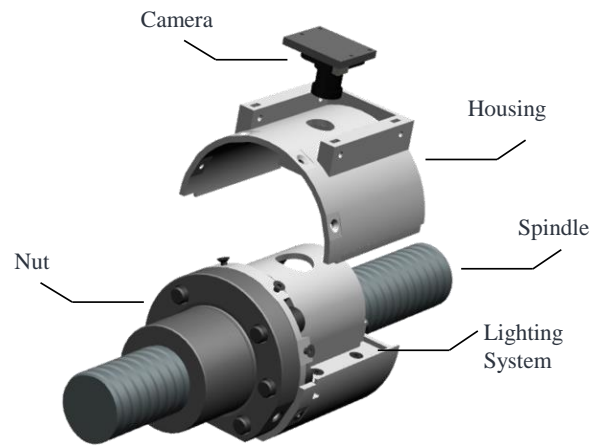

Fig. 3 Main components of the camera system used to generate the training data as described in (Schlagenhauf et al. 2019).

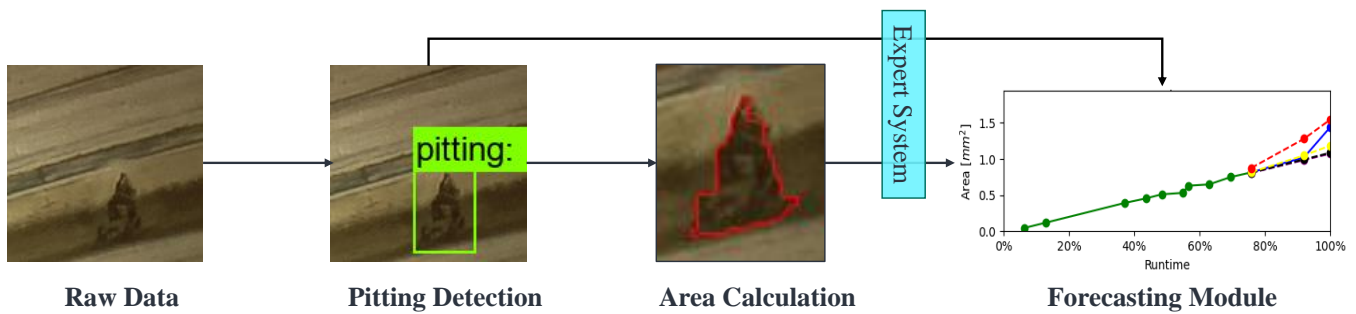

Fig. 4 Pipeline to forecast the progression of a failure. From the raw data the detection module first detects the failure in the image. This region is then passed to the area calculation module which calculates the size of the defect which is then used together with its progression history in a forecasting module to predict the future size of the defect.

Every four hours the whole spindle is scanned by the system. The experiments are undertaken with an axial load of ~14kN. The external conditions can be regarded as being similar to an industrial environment since no special protection measures were taken during the experiments and pollutions as well as lubricants are show on the spindle.The BSD-Nuts are prepared with standard wipers. Due to (Schopp 2009) changing the axial load does not influence the way a surface defect grows (the growth function is the same) but only influences the speed of growth.

The camera has a resolution of 2592x1944 Pixels and an LED lighting is used. Because of the kinematics of the BSD, the whole raceway passes the camera lens and the author's crop images of the size of 190x190 pixels automatically from the larger images. This setup can be easily adjusted to specific needs. This process is depicted together with exemplary images used for training in Fig. 5. The authors extracted in total 230 images where 60 images are put apart for testing the model and a 70/30 split of the remaining images is used for training and validation of the defect detection model. For the training of the threshold prediction model, which is described later on in more detail, the authors used 600 images from the same data source but without drawing bounding boxes around the defects. There is an intersection between the sets, but this does not influence the results since the models are distinct. These images are divided into 6 lighting categories for the threshold prediction. A 70/30 train, validation split is used for model training.

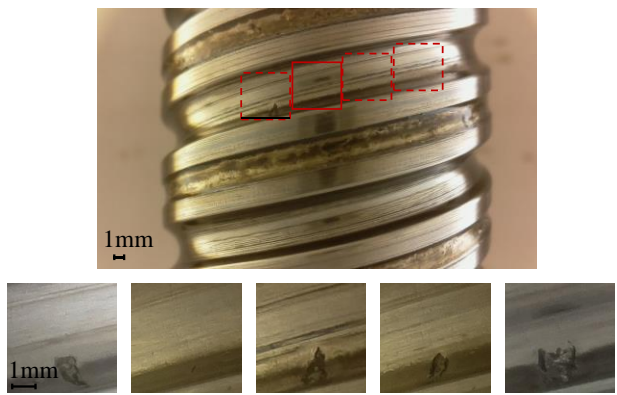

Fig. 5 Process of generating Training Data from original camera images. The images show different conditions with respect to lighting and pollution.

### B. Defect Detection

To be able to predict the evolution of the size of defects, first the defects must be located in the image. The TensorFlow Object Detection API is used to set up the object detection model. As a pre-trained model, the EfficientDet D0[1] 512x512, pre-trained on the COCO dataset, is used . EfficientDet employs EfficientNet as a backbone network which is pre-trained on the ImageNet data set. As a featured network serves the weighted bi-directional feature pyramid network (BiFPN) (Tan et al. 2020). For the object detection part, the last layer of the pre-trained model is fine-tuned on the 120 images of pittings. The convolutional base is not changed. The model is trained using a NVIDIA Tesla T4 hardware for 2000 epochs.

---

[1] For model details and further information visit the model repo: https://github.com/tensorflow/models/tree/master/research/object_detection

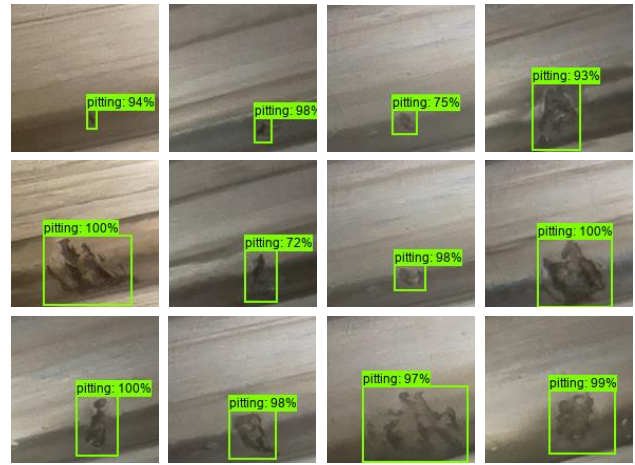

Fig. 6 Surface defects in different sizes detected by the Pitting detection model

If an image with pitting is passed through the model, the model detects the pitting and outlines the pitting with a bounding box as depicted in Fig. 6. The system yields a validation accuracy of 92%. These bounding boxes serve as Region of Interest for the following threshold model. As a result, this model could be used as a standalone model for the forecasting step in case that the approximate size of a pitting is sufficient.

By passing on only the content of the bounding box, this area then contains fewer disturbing factors, and the contour of the pitting can be determined more reliably. During the detection step, it could happen that the model cannot find an object in a specific image. In that case, it has to be differentiated if the model has detected a pitting at an earlier point in time. If this is the case, by domain knowledge, it is known that there has to be a pitting why it helps use a slightly larger bounding box as used on the same position at t-1. This is beneficial because it is not possible that the failure has shrunken or disappeared entirely. To summarize: If there is a bounding box in step t, there has to be one in step t+1. This expert behaviour based on domain knowledge is implemented by the expert system described later on. The authors provided the model with 60 additional test images containing defects of different sizes to validate the detection model. Fig. 7 shows the distribution of defects together with the information if the pitting was detected correctly (the whole pitting lies within the bounding box) or not. All pittings corrrectly detected by the object detection model are depicted in green. The defects marked in blue are the pittings for which no bounding box was found. The model accurately detects most defects and only misses some small pittings. Fig. 8 shows examples for which the model failed to detect the failures. However, since the goal is to use the bounding box as a region of interest for the next step, it is possible, as already mentioned, to use the previously found bounding box of t-1 on the same position and assume that there must be a pitting.

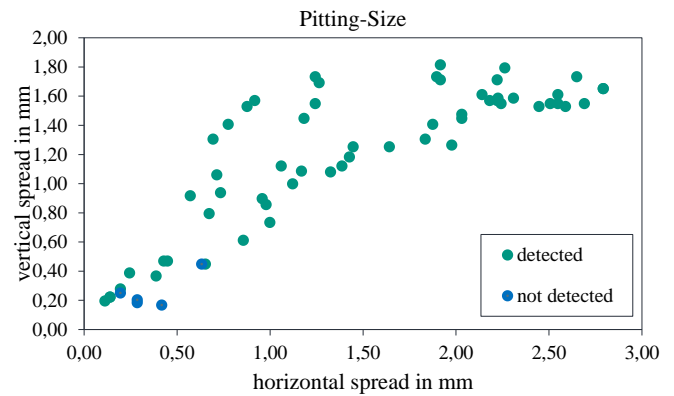

Fig. 7 Detection of Pittings of different sizes over the spread of the pittings in horizontal and vertical direction. Green means that ta pitting was detected and blue that it was not detected.

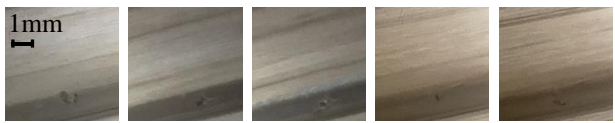

Fig. 8 Mainly very small Pittings from the test dataset not detected by the model.

*C. Threshold Prediction*

Processing the regions inside the found bounding boxes, the goal of the threshold prediction model is to find an appropriate threshold such that the failure can be extracted as precisely as possible. To extract the failure, the authors use a classical threshold-based (findcontours) approach provided by the OpenCV foundation (Bradski 2000).

The algorithm yields the best results in a binary image since there the regions are maximally distinguishable. Hence the goal is to process the image into a black and white image separating the failure area. Since of the diversity of the BSD images there is no unique appropriate global threshold value for all images. The threshold value is an integer in the range [0,…,255] marking the border below which all values are set to 0, and all values above are set to 255. Experiments show that also automatic global and local threshold finding algorithms such as the method of Otsu, which is a widely used algorithm to find the global threshold automatically (Otsu 1979) did not work properly as depicted in Fig. 9.

The method of Otsu works especially with bimodal images by choosing the threshold as a point between the modes. This is not applicable in the here presented case since the images are not bimodal and the modes are changing over images. In the here presented approach, the authors trained a CNN model to classify images into their appropriate threshold values. The basic idea of this approach is to determine the value of a parameter, which is otherwise calculated by statistical methods such as a mean value method, with the help of a CNN. This method could be used in other areas where model parameters are to be determined as well. An example could be simulations where models must be parametrized based on some raw input data. Especially when it is difficult to determine an appropriate set of values in advance and finding a value is done by trial and error, the here presented approach can aid to accelerate the process of finding appropriate parameter values. The authors used a total amount of 600 images in a 70/30 train, validation split. The images used to train the CNN model are divided into six classes for six different background conditions resulting in six threshold values. The six background conditions are labelled by the authors based on different lighting and pollution conditions which have in turn influence on the threshold. Different conditions are shown e.g. in Fig. 5 above. For example, images taken on a spindle that is already worn often have a much darker background because dirt particles, discolored lubricant and wear particles showing on the spindle. It turned out that six classes are sufficiently fine-grained, more classes does not aid the model. As threshold values, the authors defined the values 35, 40, 45, 52, 62, 72. With these thresholds, all contours could be satisfactorily represented. Fig. 10 shows the labelling process where all threshold values are applied to all images and the most sufficient threshold value is chosen.

It can be clearly seen that for each image, a different threshold value is needed for optimal contour recognition. A box surrounds the best value in each case. This process can be understood as labelling images with useful thresholds.

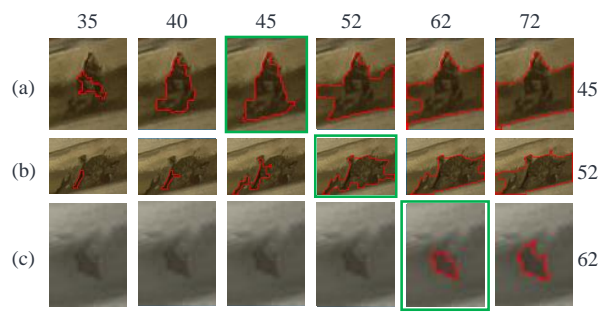

Fig. 10 Visualization of different thresholds applied to different images and comparison to the predicted threshold value (right side)

In the next step, the authors trained a CNN model by using the images together with the threshold labels. The CNN model is a manually build model with 4 convolutional layers and 2x2 max pooling operation following each convolutional layer. The convolutional base is followed by two fully connected layers with 64 neurons each. Relu activation is used in all convolutional and dense layers. The dense layers are followed by 0.1 dropout layers. The final layer consists of six neurons applying softmax activation. The used optimizer is Adam, with a learning rate of 0.0001. The training was done using NVIDIA Tesla T4 hardware for 1000 epochs.

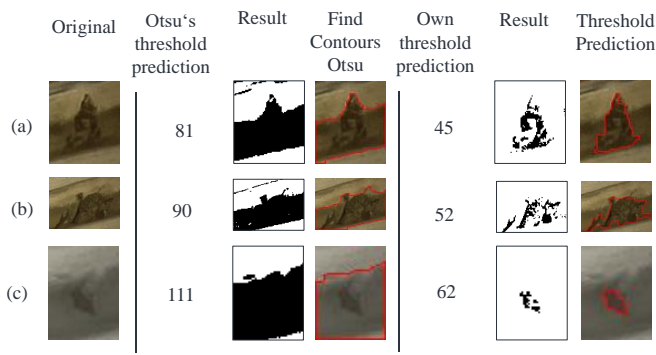

Fig. 9 Comparison between the Otsu Algorithm for finding thresholds with the proposed threshold prediction approach

The authors achieved a validation accuracy of 92% which shows that the model can accurately predict the best-suited threshold value. In advance to applying the method, the authors implemented a four-step image pre-processing to aid the segmentation results. The four conducted steps are Thresholding, Invert Bitwise, Morphological Dilation and Morphological Erosion (Fig. 11).

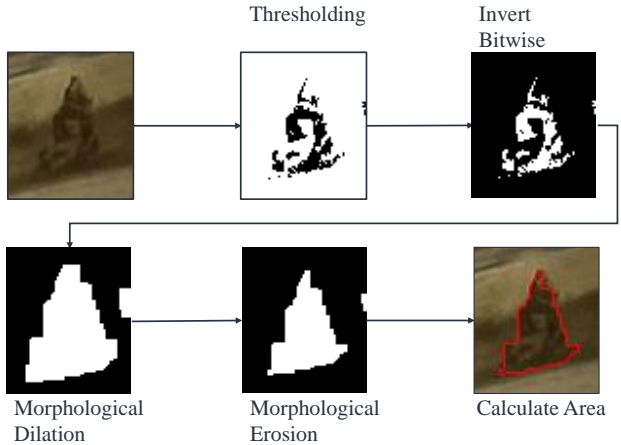

Fig. 11 Data pre-processing pipeline to calculate the defect area

### D. Expert System

Once the area of a failure is calculated, it could be used in the forecasting model to predict the expected failure size. As mentioned above, there is usually some valuable domain knowledge about the failure's visual characteristics as well as the wear mechanisms in technical domains, which could aid the machine learning approach. This is true for many areas where substantial domain knowledge is available. By intelligently combining the domain knowledge with Machine Learning systems, the intelligent systems could be improved by the human experts introducing information not easily learnable from the available data (expert system). One example is the fact that a defect on the spindle cannot shrink over time, but only grow larger. It is important to mention that the knowledge base for the expert system is implicitly available encoded in the experience of the expert and is available with zero additional data points. This means that because it is already existing, task is to properly formulate and implement the expert knowledge in an algorithm to support data driven approaches. In the here presented case, a strong characteristic of the pitting is e.g. that it has sharp corners and a somewhat darker colour than the surroundings. This knowledge was not explicitly formulated above but implicitly used by the contour finding algorithm to find the borders of the failures. With the above described steps, it is possible to measure the size of a pitting very accurately, but due to oil or pollution on the ball screw, the appearance of the pitting area can vary. In these cases, the calculated pitting area would deviate from reality. Fig. 12 opposes the predicted progression of pitting (Quantification) to the results implementing the expert system and the ground truth data over 28 timesteps after the first pitting has been observed. The y-axis represents the pitting's size in connection with the respective time steps (x-axis). The blue data series reflects the ground truth data whilst the grey data series represents the values measured by the model without the implementation of the expert system. The green line represents the results after implementing the expert system. To make the prediction process more clear, predictions are made in a successive manner like it would be applied in reality during different steps in time. For each step, a linear regression model (purple line) trained on the data already processed by the expert system is added.

It can be seen how the model without expert system (grey) matches the ground truth data (blue) at the beginning. However, there are some strong outliers in later time steps. Additionally, it can be seen that the model partly overestimates the size of pitting in later time steps and also predicts smaller sizes for later timesteps. Though by domain knowledge, it is known that this effect of decreasing pitting sizes is not possible. Hence, either the model has overestimated the size of a pitting in earlier timesteps or underestimates its size. To address this issue, domain knowledge is introduced by applying a two-step algorithm represented by the green line. As a baseline for the expert system, three possible cases are discussed.

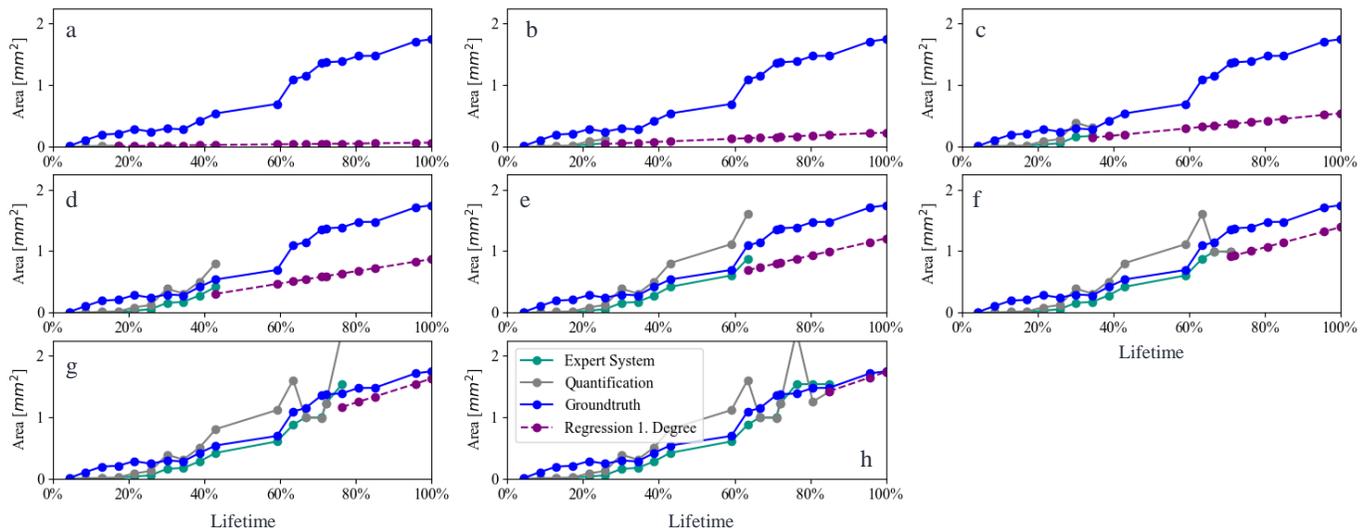

Fig. 12 Comparison of ground truth failure progression to the prediction of the applied expert system. Progression from a to h.

First, the area in time step t+1 is slightly larger than the area in time step t. In this case, the proposed value is valid. Second, the area in time step t+1 is disproportionately large which may cause from severe pollutions on the surface. In this case, the average of the defect size in time step t+1 and the defect size in time steps t and t-1 is calculated and used as the predicted size. In this way, rough outliers are averaged which yields a smoother curve. Third, the area in t+1 is smaller than in time step t, which is impossible. In this case, the pitting size at t+1 is predicted as the size in t and hence remains the same.

Therefore, the expert system as the last step of the pipeline compensates for erroneous measurements in the previous steps and results in a more accurate prediction model which shows that implementing the expert system aids the prediction.

### E. Forecasting Function

The linear regression fitted to the data above seems to fit the data well. To double-check this assumption, the authors fitted a set of functions to the data to measure their root mean squared error (RMSE) between the predictions and the ground truth data. As functions the authors chose the linear regression, a second and a third order polynomial and an exponential function. As a base method for determining the forecasting quality, the classical RMSE is used with:

$E = \frac{1}{J}\sum_{j=1}^{J}\sqrt{(\hat{a}_x^{t+j} - a_x^{t+j})^2}$ where $\hat{a}_x^{t+j}$ is the size predicted by the expert system $j$ timesteps ahead of $t$ and $a_x^{t+j}$ is the ground truth value at the same time step. Hence the sum of the distances between all predicted and ground truth points is measured. The closer the predicted values match the true values, the better the function fits the data.

The issue here is that this only gives a measure on already observed points which is not appropriate in practice since the higher the polynomial the better the function will fit the data. A polynomial with degree *n* can match *n* data points with *E=0* though this function will probably not be a good estimator for future points. The goal is not to have a function which is precise on already observed points but on future points. Hence the model is created on points up to timestep *t* and the prediction precision is measured on all points *j* in the future.

From a practical point of view, predictions that are correct in the very near future and predictions that are correct in the very far future are less important than middle-term predictions. This is the case because to maintain a component in time, it is necessary to look some time ahead, which covers the needed time for the preparations for maintenance. Hence, it is of little value to have a model which is very accurate for the very near future but fails in the middle and long term future since then the time to plan the maintenance is not sufficient. The same is true for predictions very far in the future because the preparations for maintenance take some time *p* and if the prediction horizon is much greater than *p* this is of little extra value since it will not change the planning behaviour. Hence, there is a middle-term "sweet spot" in which a model should be as accurate as possible. The selection of this "sweet spot" horizon changes for different companies and processes. To implement this behaviour the formulation of the RMSE is extended to incorporate a time component:

$E_\alpha = \frac{1}{J}\sum_{j=1}^{J}\sqrt{f(j)(\hat{a}_x^{t+j} - a_x^{t+j})^2}$ where $\alpha$ is a parameter for how far in the future the highest attention is laid. $\alpha$ should be odd for mathematical convenience. The authors chose $\alpha$ as 7. The function $f(j)$ is chosen as a bell-shaped function with $f(j) = e^{-0.15*\left(\left(ceil\left(\frac{\alpha}{2}\right)-j\right)^2\right)}$ where $ceil(.)$ means that the resulting float is rounded up to the next integer which is 4 in that case. The function $f(j)$ is symmetric and takes its maximum value of 1 at $j = 4$ and smaller values for $j > ceil\left(\frac{\alpha}{2}\right)$ and $j < ceil\left(\frac{\alpha}{2}\right)$. Hence the resulting RMSE is weighted and the value which lays 4 time steps in the future receives the highest weight.

Additionally, from a practical point of view a function is wanted that is as data efficient as possible. The function should fulfil the above criteria with as little data as possible. To implement this behaviour, the Loss Function is further developed to incorporate the reciprocal of the amount of data points used for training. The model is trained with progressively more data points where the minimum number of available data points is set to four, until the calculation of the error is started. The RMSE is then summed over all $RMSE_\beta$ where $\beta$ are the number of data points available for training. The final loss term is:

$$E = \sum_{\beta}^{N}\beta^{-1}\left(\frac{1}{J}\sum_{j=1}^{J}\sqrt{f(j)(\hat{a}_x^{t+j} - a_x^{t+j})^2}\right) = \sum_{\beta}^{N}\beta^{-1}RMSE_\beta^\alpha$$

This final error term is used to compare the functions. The result is shown in Fig. 13. The bar-plot shows that the assumption that the linear function yields the best results can be validated. The linear function closely matches the data which is plausible because by inspection the data follows a linear function of order one. The predicted curves indicate that all higher order polynomials fail to match the data because they are too oscillatory. The linear model is chosen as the appropriate model to fit the evolution in size of a defect.

### IV. VALIDATION

To validate the approach, the authors test the model on a new failure to check if first, the detection model is together with the expert system, able to accurately quantify the size of a failure. Secondly, the authors check the assumption that a linear function is able to forecast the size of a failure on a BSD. Additionally, the authors added a symbolic, industry-dependent measure for the allowed wear as well as a +-20% confidence band around the ground truth data. As allowed lifetime, the authors choose an area of 0.9 mm² as the industry dependent lifetime of a component. This border has to be adjusted depending on the industry but is chosen such that there is sufficient time until the mechanical breakdown of the component. The results are shown in Fig. 14.

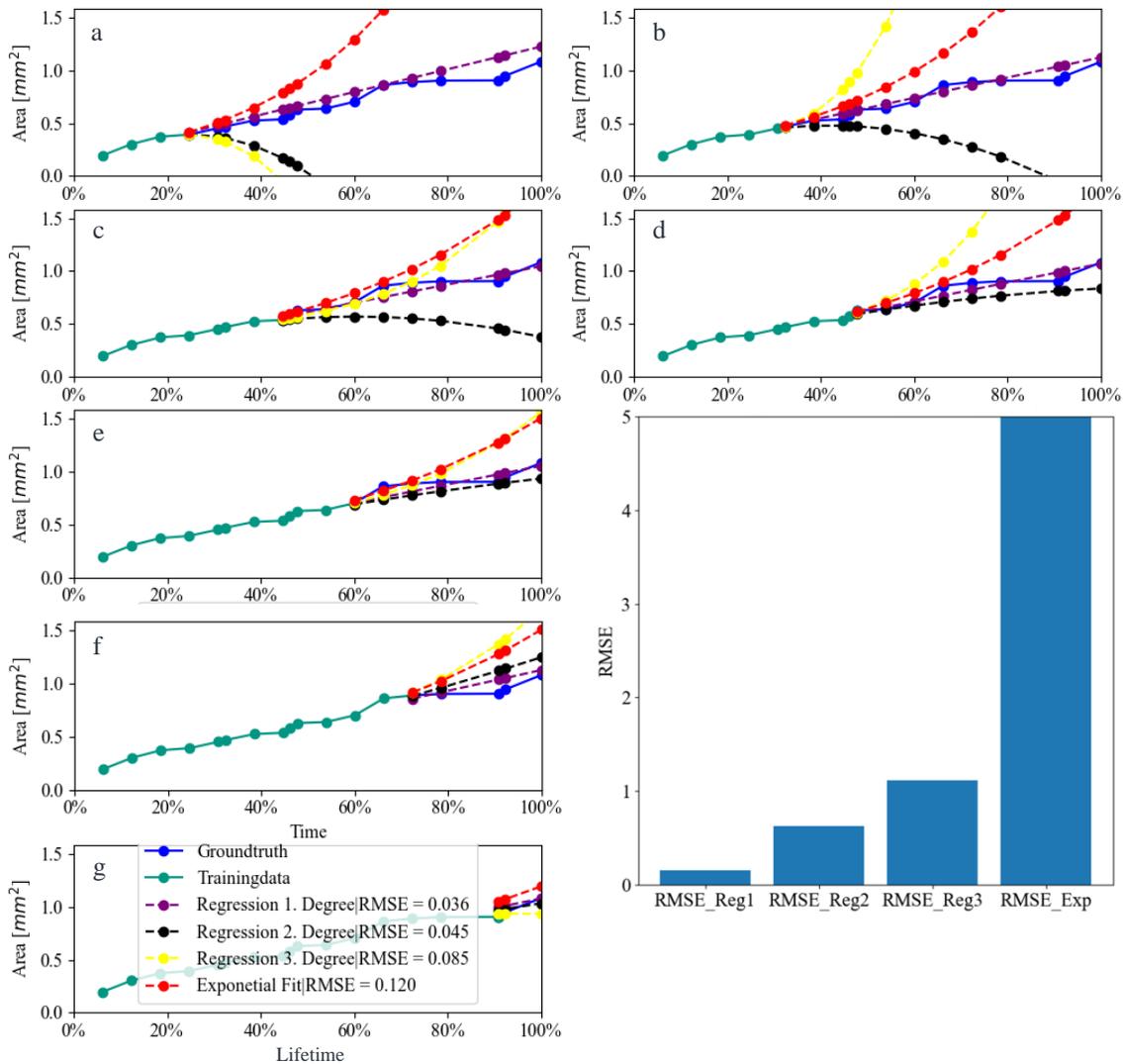

Fig. 13 RMSE for several functions in comparison. Bar plot shows the comparison of the RMSE values for different functions.

Here it is important that the green line represents the data extracted by the expert system up to a specific time step. The forecasting model is built on this data. It is notable that the model is accurately working and is able to predict the failure in advance. During the first time steps, the system is underestimating the size of the failure which is why the regression line is flat. At about 20% of the lifetime, the system correctly predicts the size of the defect and the regression line matches the ground truth data quite well which yields the ability to predict the end of life of the component after 40% of the lifetime. In later time steps, the model over-estimated the size of the failures which could result from, e.g. pollutions on the spindle. This led the system to underestimate the remaining lifetime of the component slightly which is in practice a not severe issue since this only would lead the maintenance to take action a bit earlier and give kind of an additional reserve. Additionally, the authors added images of the failure progression for specific time steps. This indicates the functionality of the model and shows the growth of a defect on the spindle. At the top the detection model finds the pitting where at the bottom the size quantification module quantifies the accurate area of the failure.

V. CONCLUSION

The authors presented an end to end pipeline for the detection and the forecasting of failures on BSD into the future. The model can be easily applied to other surface defects not only on machine tool components since the underlying methodology is transferable. The presented system can be used by practitioners as a predictive maintenance tool to plan maintenance operations in time before component failure. Due to the incorporation of domain knowledge, the authors improved the prediction system, which accurately detected failures but misinterpreted the severity of some failures due to pollution on the spindle. The authors noticed this effect even during the expert based labelling of the ground truth data, with the difference that the human expert unconsciously incorporates domain knowledge into the labelling process. This limitation is obvious, especially for early failures where the whole failure comprises only a few pixels. A practical though cost-increasing improvement of the system would be to use a camera with a higher resolution which possibly could reduce the described effect.
The model can be used in three setups. The first setup is to use the image data and label the size of the failures manually by experts and then only use the forecasting model to predict the size of the failures into the future. This is robust in terms of the extraction of the failure size but needs manual efforts. The second approach is to use the whole pipeline but omit the size extraction module and only use the failure detection module. This can be done if an exact pixel resolution of a failure is not important. The third way is to use the whole pipeline.

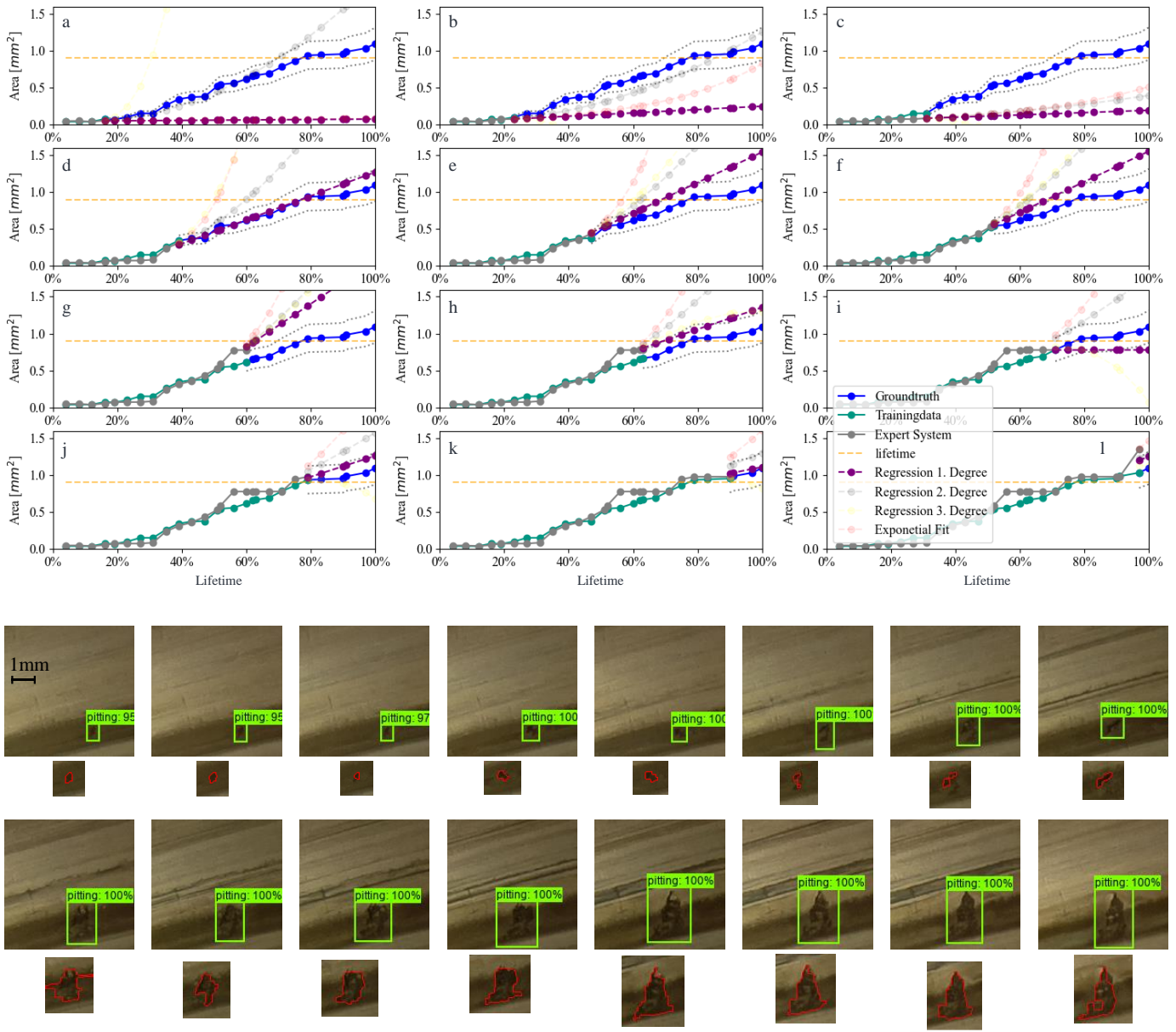

Fig. 14 Validation of the failure forecasting system: Top: Forecasting model based on the extracted sizes of failures. Bottom: Evolution of the failure used for validation where the detected pittings in the images are framed by the detection module and their size is extracted by the size quantification module.

Additional expert queries could be implemented to make the model more robust. For instance, by asking an expert if a failure has been correctly detected after a failure has been detected by the system. This could exclude false positives. Additionally, this information could be used in further research to re-train and adapt the model continuously. In additional further research, it should be proposed to automatically learn the domain knowledge induced by the human expert by using models that also incorporate time as a feature and therewith learn that failures are growing over time. Heavy pollutions which cover the valuable regions in an image lead to a natural limitation of vision-based systems. This limitation can be interpreted as noise in the pixel space. If the noise is too prevalent, the necessary information is no longer available. Transferred to signals in the time domain, pollutions act like strong inferring signals. The significant advantage of a vision-based approach is that the noise can be easily identified by inspection. Elaborating on this, a useful extension to the vision-based approach would be the combination with time domain signals which are less influenced by pollutions. Possible signals could be acoustic emission or the motor current. Here the motor current is the suggested first choice of the authors since it does not need additional sensors. By training a motor current based system using the image data, the system should be able to connect the motor current signal to images showing defects and therewith learns how the motor signal (or acoustic emission signal) of a defect looks like in an image-based way. The Hypothesis is that the motor current signal is less affected by covering pollutions and hence as a tandem even more accurate predictions could be made. As open question remains if the signal of the motor current is affected by small defects. This idea can easily be extended to multiple sensor systems where the image data provide mutual information. Student-Teacher approaches are thinkable where the knowledge of multiple sensor systems is incorporated into one intelligent model.

## VI. DECLARATIONS

### A. Funding

This work was supported by the German Research Foundation (DFG) under Grant FL 197/77-1.

### B. Conflicts of interest/Competing interests

See attached conflicts of interest file. There are no competing interests.

### C. Availability of data and material

Not applicable.

### D. Code availability

The code is available at https://github.com/2Obe/Pitting_Pred_Maintenance

### E. Authors' contributions

Tobias Schlagenhauf: Concept, Writing, Coding.
Niklas Burkhardt: Writing, Coding, Review

### F. Ethics approval

Not applicable.

### G. Consent to participate

Not applicable.

### H. Consent for publication

Not applicable.